\documentclass[twoside]{article}

\usepackage[accepted]{aistats2025}
\usepackage{graphicx}
\usepackage{amsmath}
\usepackage{amssymb}
\usepackage{booktabs}
\usepackage{cite} 
\usepackage{placeins}
\usepackage{lscape}
\usepackage{longtable}
\usepackage{multirow}

\usepackage[capitalize]{cleveref}
\crefname{section}{Sec.}{Secs.}
\Crefname{section}{Section}{Sections}
\Crefname{table}{Table}{Tables}
\crefname{table}{Tab.}{Tabs.}

\usepackage{algorithm}
\usepackage{algorithmic}
\usepackage{amsmath}
\usepackage{amssymb}
\usepackage{mathtools}
\usepackage{amsthm}
\theoremstyle{plain}
\newtheorem{theorem}{Theorem}[section]
\newtheorem{proposition}[theorem]{Proposition}

\theoremstyle{definition}

\theoremstyle{remark}

\usepackage{subcaption}
\usepackage[round]{natbib}
\usepackage{url}
\usepackage[hyphenbreaks]{breakurl}

\begin{document}

%

%

\twocolumn[

\aistatstitle{\underline{S}eparate, Dyna\underline{m}ic \underline{a}nd Diffe\underline{r}en\underline{t}iable (SMART) Pruner for Block Pruning on Computer Vision Tasks}

\aistatsauthor{Guanhua Ding\footnotemark[1] \And Zexi Ye\footnotemark[1] \And Zhen Zhong\footnotemark[1]\footnotemark[2] \And Gang Li\footnotemark[1] \And David Shao\footnotemark[1]}

]
\footnotetext[1]{Affiliation: Black Sesame Technology}
\footnotetext[2]{Corresponding author: Zhen Zhong (\texttt{zhen.zhong@bst.ai})}



\begin{abstract}

Block pruning, which eliminates contiguous blocks of weights, is a structural pruning method that can significantly enhance the performance of neural processing units (NPUs). In industrial applications, an ideal block pruning algorithm should meet three key requirements: (1) maintain high accuracy across diverse models and tasks, as machine learning deployments on edge devices are typically accuracy-critical; (2) offer precise control over resource constraints to facilitate user adoption; and (3) provide convergence guarantees to prevent performance instability. However, to the best of our knowledge, no existing block pruning algorithm satisfies all three requirements simultaneously. In this paper, we introduce SMART (Separate, Dynamic, and Differentiable) pruning, a novel algorithm designed to address this gap.  SMART leverages both weight and activation information to enhance accuracy, employs a differentiable top-k operator for precise control of resource constraints, and offers convergence guarantees under mild conditions.  Extensive experiments involving seven models, four datasets, three different block types, and three computer vision tasks demonstrate that SMART pruning achieves state-of-the-art performance in block pruning.

\end{abstract}

\section{Introduction}
\label{sec:intro}


Block pruning\citep{siswanto2021reconciling}, which prunes out contiguous blocks of weights, is a structural pruning method that can significantly boost the performance of NPUs\citep{google_tpu_blog}. NPUs are specialized hardware designed to achieve peak performance in neural network inference tasks and widely used in edge inference applications. The primary advantage of block pruning lies in its alignment with the intrinsic hardware architecture of NPUs, which are specifically designed to accelerate deep learning inference tasks. NPUs rely on Multiply-Accumulate (MAC) arrays for computation, where each cycle involves matrix multiplication of sub-blocks of weights and data. By pruning certain sub-blocks of weights, corresponding computational cycles can be skipped, thereby achieving real acceleration with low overhead \citep{d2024weight}. Driven by the real acceleration needs of a leading autonomous driving chip company, our objective is to develop industrially applicable, state-of-the-art block pruning algorithms.

In industrial applications, an ideal block pruning algorithm should satisfy three critical criteria. First, it should maintain high accuracy across a broad spectrum of models, tasks, and datasets. This is imperative because pruning algorithms are often developed as third-party tools \citep{intel_neural_compressor_repo, nvidia_apex, qualcomm_aimet, apple_coremltools}, with developers typically lacking access to end-users' models and data. Moreover, many real-world applications are highly sensitive to accuracy; for example, in autonomous driving systems, even minor performance degradations can have significant safety implications. Second, the algorithm should provide precise control over resource constraints, such as sparsity levels and multiply-accumulate (MAC) counts. This capability enables users to easily apply the method to obtain baseline results with minimal effort. If the initial outcomes are promising, they can then invest additional resources to fine-tune the algorithm for optimal performance tailored to their specific requirements. Third, although not mandatory, offering a theoretical convergence guarantee is highly desirable. In scenarios where algorithm developers lack knowledge of the user's specific model, dataset, or task, the absence of convergence guarantee can lead to performance instability.

To address these limitations, we propose the SMART pruning algorithm. Our approach reformulates the original pruning problem---an $L_0$-norm constrained optimization problem---into an unconstrained optimization problem by employing masks generated using a standard Top-$k$ operator. We then relax this operator into its differentiable variant, which allows us to apply standard stochastic gradient descent (SGD) for optimization. To reduce the approximation gap between the differentiable and standard Top-$k$ operators, we dynamically decrease the temperature parameter; as the temperature approaches zero, the differentiable Top-$k$ operator converges to the standard Top-$k$ operator. In our algorithm, we utilize information from both weights and activations to enhance pruning performance, while the differentiable Top-$k$ operator ensures precise control over resource constraints. Additionally, we establish the conditions for convergence of our algorithm, demonstrating that convergence is achieved when the temperature parameter decays sufficiently fast. With these core components, our algorithm satisfies industrial requirements and achieves state-of-the-art performance across various computer vision tasks and models.

The contributions of this paper are as follows:
\begin{itemize}
    \item We introduce a novel SMART pruning algorithm, specifically designed for block pruning applications, that successfully tackles three critical challenges prevalent in the industry.
    \item We analyze how a dynamic temperature parameter aids in escaping non-sparse local minima during training and establish the conditions for convergence guarantees.
    \item Our theoretical analysis demonstrates that as the temperature parameter approaches zero, the global optimum solution of the SMART pruner converges to the global optimum solution of the fundamental $L_0$-norm pruning problem.
    \item Experimental studies show that the SMART pruner achieves state-of-the-art performance across a variety of models and tasks in block pruning applications.
\end{itemize}

\section{Related Works}

Numerous studies have been conducted to improve neural network pruning algorithms. The most intuitive approach is criteria-based pruning, such as magnitude pruning \citep{reed1993pruning}, weight and activation pruning \citep{sun2023simple}, attention-based pruning \citep{zhao2023automatic}, and SNIP \citep{lee2018snip}. These methods operate on the assumption that weights can be ranked based on a specific criterion, with the least significant weights being pruned accordingly. The advantage of this approach is its stability and simplicity. However, the most suitable criterion varies as the weight and activation distributions change. It is challenging to determine the most appropriate criterion for a given pruning problem without empirical testing \citep{he2020learning}.

To alleviate this problem, many research efforts incorporate weight importance ranking into the training or inference process. One of the most straightforward ways of doing this is through black-box optimization. For example, AutoML for Model Compression (AMC) \citep{he2018amc} employs reinforcement learning to create pruning masks, leveraging direct feedback on accuracy from a pruned pre-trained model. AutoCompress \citep{liu2020autocompress} uses simulated annealing to explore the learning space, while Exploration and Estimation for Model Compression adopts Monte Carlo Markov Chain (MCMC) \citep{zhang2021exploration} methods to rank weight importance based on the direct feedback of the pre-trained model. The major drawback of black-box training-based pruning algorithms is that their convergence properties are not as efficient as gradient-based methods \citep{ning2020dsa}.

Even though numerous works have been conducted on gradient-based methods, they are mainly developed for N:M or channel pruning, which cannot be directly applied to block pruning applications. For example, the Parameter-free Differentiable Pruner (PDP) \citep{cho2024pdp} uses weights to construct within-layer soft probability masks and updates the weights using back-propagation. Learning Filter Pruning (LFP) \citep{he2020learning} determines the most suitable criteria for each layer by updating the importance parameter of each criterion through gradients. Dynamic Network Surgery (DNS) \citep{guo2016dynamic}, Global and Dynamic Filter Pruning (GDFP) \citep{lin2018accelerating}, and Dynamic Pruning with Feedback (DPF) \citep{lin2020dynamic} use hard binary masks calculated from weights to achieve sparsity for each layer, updating the weights using straight-through estimator (STE) methods. Discrete Model Compression (DMC) \citep{gao2020discrete} applies separate discrete gates to identify and prune less important weights, with gate parameters updated through STE methods. However, these methods lack cross-layer weight importance ranking components, making them less suitable for block pruning applications. Conversely, the CHEX pruning algorithm \citep{hou2022chex} ranks within-layer importance through singular value decomposition of each channel's weights, and Differentiable Markov Channel Pruning (DMCP) \citep{guo2020dmcp} assumes within-layer importance can be ranked sequentially. However, these methods operate under the assumption that within-layer output channels share the same input features, making them not directly applicable to block pruning applications as well.

Some gradient-based methods can simultaneously learn both cross-layer and within-layer importance rankings, yielding promising accuracy results across various models, tasks, and datasets. However, they often require extensive parameter tuning or additional training flows to control resource constraints, making them less appealing for industrial applications. For example, Pruning as Searching (PaS) \citep{li2022pruning} and Soft Channel Pruning \citep{kang2020operation} can learn both cross-layer and within-layer importance during training. However, they rely on penalty terms to control target sparsity, which requires significant tuning effort in real applications. Differentiable Sparsity Allocation (DSA) \citep{ning2020dsa} can learn cross-layer importance through gradients and rank within-layer importance using predefined criteria. ADMM is then adopted to handle resource constraints. However, ADMM requires reaching at least a local optimum for each iteration to ensure convergence \citep{boyd2011distributed}, significantly increasing training costs. In practical applications, ADMM iterations are usually switched before reaching a local optimum, which can lead to a lack of convergence guarantees and instability in the training process.

Based on the foregoing analysis, no existing methods can simultaneously learn both cross-layer and within-layer weight importance rankings through gradients, which are essential for achieving high accuracy, precise control of resource constraints, and ensuring algorithm convergence. Our SMART pruning algorithm bridges this gap, advancing the industrial frontier of block pruning.

\section{Methodology}

In this section, we will introduce the methodology of SMART pruner. Specifically, in section 3.1, we will introduce our differentiable Top $k$ operator, which is the cornerstone of our SMART pruner. Section 3.2 elaborates on the SMART pruning algorithm, including the problem formulation and detailed training flow. Finally, section 3.3  elaborates on our dynamic temperature parameter trick, explaining its role in avoiding non-sparse local minima and the conditions for convergence guarantee.

\begin{figure*}
\centering
\includegraphics[width=0.8\textwidth]{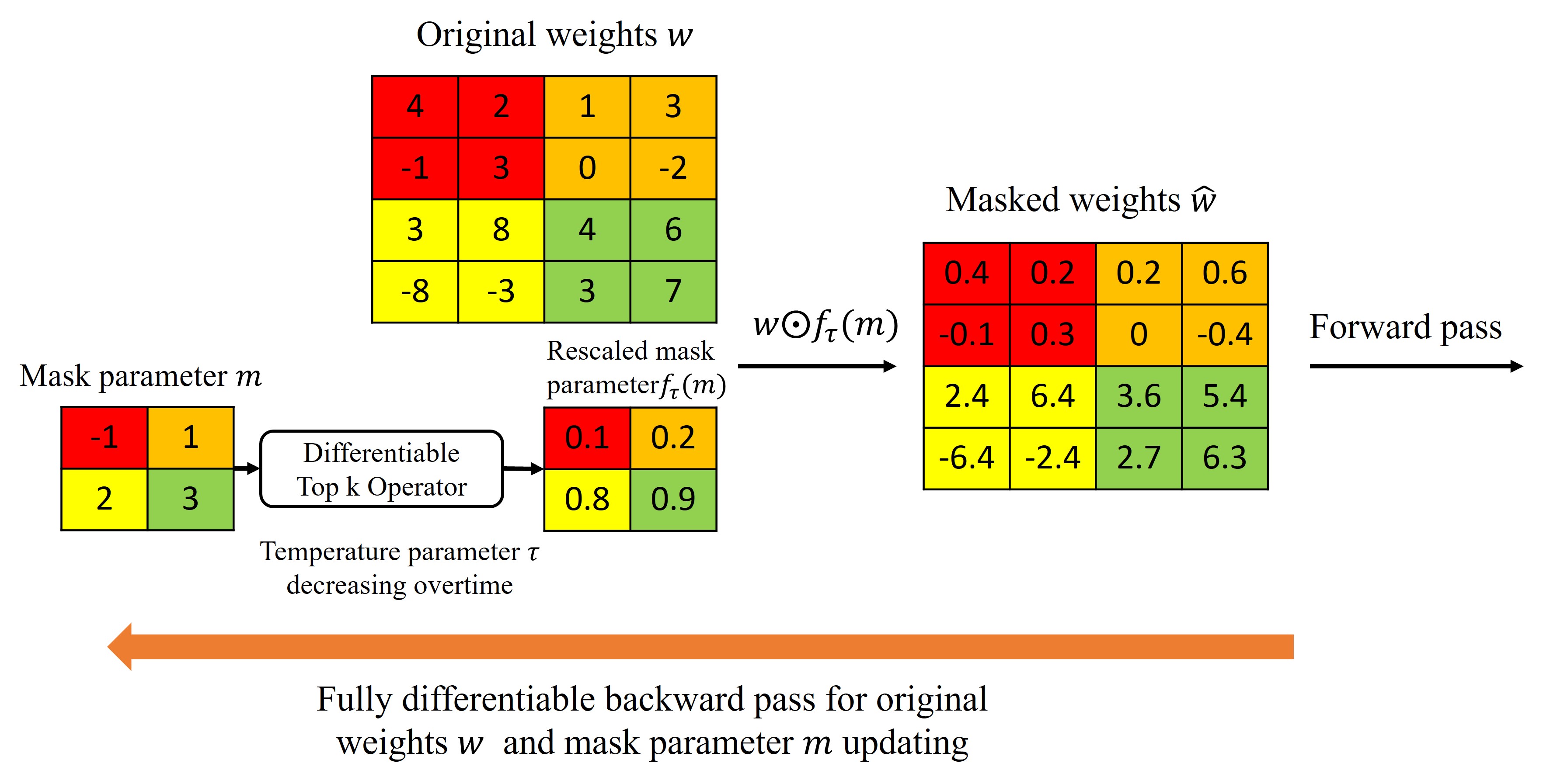} 
\caption{Illustration of the SMART pruner. In the forward pass, the original weights (top left) are element-wise multiplied by a rescaled importance mask, $f_\tau(m)$, generated from a differentiable top-$k$ operator, to obtain the masked weights matrix (top right). In the backward pass, the original weights parameter $w$ and mask parameter $m$ are updated via back-propagation.}
\label{fig:figure1} 
\end{figure*}

\subsection{Differentiable Top $k$ Operator}
In the deep learning field, the selection of the Top $k$ elements is a fundamental operation with extensive applications, ranging from recommendation systems to computer vision \citep{shazeer2017outrageously,fedus2022review}. Standard Top $k$ operators, however, suffer from non-differentiability, which impedes their direct utilization within gradient-based learning frameworks. Typically, the standard Top $k$ operator $Top_k (x_i)$ could be written as:
\begin{equation}
Top_k(x_i) = 
\begin{cases} 
1 & \text{if } x_i > x_{\rho_{N-k}} \\
0 & \text{if } x_i \leq x_{\rho_{N-k}}
\end{cases}
\end{equation}
where $x_i$ represents $i$-th input value, $\rho$ refers to the sorted permutations, i.e., $x_{\rho_1} < x_{\rho_2} < \cdots < x_{\rho_N}$.

To directly incorporate the Top-$k$ operator into the training process, numerous works have focused on designing differentiable Top-$k$ operators \citep{xie2020differentiable, sander2023fast}. Our requirement of differentiable Top $k$ operators is (i) simple to implement and (ii) capable of arbitrary precision approximation. To achieve this goal, we modify the sigmoid-based Top $k$ operator \citep{ahle2023differentiable} by introducing the temperature parameter and the mathematical definition of our differentiable Top $k$ operator is as follows:
\begin{gather}
f_{\tau,i}(\mathbf{x}) = \sigma\left(\frac{x_i}{\tau} + t(x_1, \ldots, x_N)\right) \\
\text{\centering subject to}
\sum_{i=1}^{N} f_{\tau,i}(\mathbf{x}) = k \notag
\end{gather}
where $N$ denotes the total number of inputs, $k$ specifies the number of largest elements to be selected by the Top $k$ operator, $\tau(>0)$ stands for the temperature parameter, and $x_i$ represents the $i$-th input variable. The function $\sigma(x)=1/(1+e^{-x})$ represents the sigmoidal activation function, mapping the input into the $(0, 1)$ interval. The determination of $t(x_1,…,x_N )$ hinges on the monotonicity of the sigmoid function. A viable approach to calculate $t(x_1,…,x_N) $ is the binary search algorithm, ensuring that the constraint of Equation (2) is satisfied.
\begin{theorem}
Suppose \( N \geq 2 \) and \( 1 \leq k \leq N \). As the temperature parameter \( \tau \) approaches zero, we have: $\lim_{\tau \to 0} f_{\tau, \rho_{N-k}}(\mathbf{x}) = \sigma \left( \frac{x_{\rho_{N-k}}}{\tau} + t(x_1, \ldots, x_N) \right) = 0, \quad \forall i \in [1, N-k]$
and $\lim_{\tau \to 0} f_{\tau, \rho_{N-k+1}}(\mathbf{x}) = \sigma \left( \frac{x_{\rho_{N-k+1}}}{\tau} \right) = 1, \quad \forall i \in [N-k+1, N]$, where \( \sigma \) is the sigmoid function.
\end{theorem}

The proof can be found in Appendix A. From Theorem 2.1, we prove that this differentiable Top $k$ operator $f_{\tau,i} (x)$ could approximate the standard Top $k$ operator $Top_k(x_i)$ to an arbitrary precision by simply reducing the temperature parameter. 

\begin{proposition}
The gradient of \( f_{\tau,i}(\mathbf{x}) \) is:
\begin{align*}
\frac{d f_{\tau,i}(\mathbf{x})}{d x_j} = \frac{1}{\tau} \sigma' \left( \frac{x_i}{\tau} + t(\mathbf{x}) \right) \left( I_{i=j} - \frac{\sigma' \left( \frac{x_j}{\tau} + t(\mathbf{x}) \right)}{\sum_{i=1}^{N} \sigma' \left( \frac{x_i}{\tau} + t(\mathbf{x}) \right)} \right)
\end{align*}
where \( \sigma' \left( \frac{x_i}{\tau} + t(\mathbf{x}) \right) \) is the gradient of the sigmoid function. The notation \( I_{i=j} \) is a conditional expression, where it equals 1 if \( i = j \), and 0 otherwise. If we further define \( \mathbf{v} = \left[ \sigma' \left( \frac{x_1}{\tau} + t(\mathbf{x}) \right), \ldots, \sigma' \left( \frac{x_N}{\tau} + t(\mathbf{x}) \right) \right]^T \), the Jacobian of our differentiable Top-k operator at \( \mathbf{x} \) is:$J_{\text{Top}_k}(\mathbf{x}) = \frac{1}{\tau} \left( \text{diag}(\mathbf{v}) - \frac{\mathbf{v} \mathbf{v}^T}{\| \mathbf{v} \|_1} \right)$.
\end{proposition}
The proof of proposition 3.2 can be found in Appendix B. From Proposition 3.2, the backpropagation of this differentiable Top $k$ operator could be easily computed. From the above derivation, it is evident that the time and space complexities of both the forward and backward passes are $O(n)$. 
\subsection{SMART pruning algorithm}
Let $L$ represent the loss function, $w$ denote the weights blocks (grouped by the pruning structure), $n(w)$ represent the total number of weights blocks, and $r$ indicate the sparsity ratio. A typical pruning problem can be formulated as solving the following optimization problem: 
\begingroup
\setlength{\abovedisplayskip}{5pt} 
\setlength{\belowdisplayskip}{5pt} 
\setlength{\abovedisplayshortskip}{5pt}
\setlength{\belowdisplayshortskip}{5pt}

\begin{align}
& \min_{w} L(w) \\
 \text{subject to } & \|w\|_0 = (1 - r) \times n(w) \nonumber
\end{align}
\endgroup

where $\|w\|_0$ denotes the zero-norm, which counts the total number of non-zero weight blocks.

\begin{theorem}
Suppose the pair \((w^*, m^*)\) is the global optimum solution of the following problem, then \( w^* \odot \text{Top}_k (m^*) \) is also the global optimum solution of the original problem.

\begingroup
\setlength{\abovedisplayskip}{5pt} 
\setlength{\belowdisplayskip}{5pt} 
\setlength{\abovedisplayshortskip}{5pt}
\setlength{\belowdisplayshortskip}{5pt}

\begin{align}
& \min_{w,m} L(w \odot \text{Top}_k (m)) \\
  \text{subj} & \text{ect to }  k = (1 - r) \times n(w) \nonumber
\end{align}
\endgroup

where \(\odot\) denotes the Hadamard product.
\end{theorem}

The proof of Theorem 3.3 can be found in Appendix C. Theorem 3.3 enables us to shift our focus from solving the problem (3) to solving the problem (4). A significant hurdle in optimizing problem (4) is the non-differentiability of its objective function. To overcome this challenge, we replace the standard Top $k$ operator with our differentiable Top $k$ operator. We define $f_\tau(m)$ such that the $i$-th element multiplication $[w \odot f_\tau(m)]_i=w_i \odot f_{\tau,i}(m)$. Then, the problem (4) is transformed into the new problem (5), which is the problem formulation for our SMART pruner:

\begin{gather}
 \min_{w,m} L(w \odot f_\tau(m)) \\
 \text{subject to } f_{\tau,i}(m) = \sigma \left( \frac{m_i}{\tau} + t(m_1, \ldots, m_{n(w)}) \right), \nonumber \\
 \sum_{i=1}^{n(w)} f_{\tau,i}(m) = (1 - r) \times n(w) \nonumber
\end{gather}
\begin{theorem}
Suppose the pair \((w^*, m^*)\) is the global optimum solution of problem (4) and \(L\) is a Lipschitz continuous loss function. Then, for any given solution pair \((\tilde{w}, \tilde{m})\) satisfying \(L(\tilde{w} \odot \text{Top}_k (\tilde{m})) > L(w^* \odot \text{Top}_k (m^*))\), there exists a \(\tilde{\tau}\) such that for any \(\tau \in (0, \tilde{\tau}]\), the inequality holds: \(L(w^* \odot f_\tau(m^*)) < L(\tilde{w} \odot f_\tau(\tilde{m}))\).
\end{theorem}
The proof of Theorem 3.4 can be found in Appendix D. From the Theorem 3.3 and Theorem 3.4, our SMART pruner inherently strives to solve the fundamental pruning problem directly as temperature parameter approaches 0. By directly solving the fundamental pruning problem, our SMART algorithm mitigates the impact of regularization bias, resulting in superior performance over existing arts. During the training procedure, our SMART pruner employs projected stochastic gradient descent to solve the problem (5). The detailed training flow of our SMART pruning algorithm is summarized in Algorithm 1.

\begin{figure}
\centering
\includegraphics[width=0.5\textwidth]{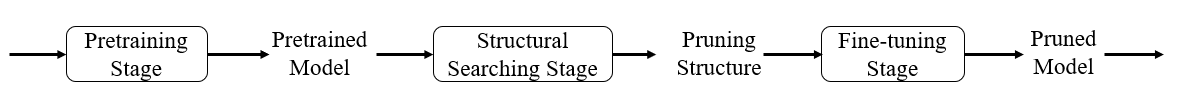} 
\caption{Training Flow of SMART Pruner. This illustrates a three-stage process: pretraining to develop the initial model, structural searching to determine the pruning structure, and fine-tuning to enhance performance post-pruning.}
\label{fig:figure2} 
\end{figure}

\begin{algorithm}[ht]
\caption{Training flow of SMART pruner for block/output channel pruning.}
\begin{algorithmic}
\STATE $\textbf{Input:} \ s, l, r, \tau_s, \tau_e, k, w = [w_0, \ldots, w_{n(w)}]$.
\FOR{$e \in [0, 1, 2, \ldots, s]$}
    \FOR{each mini-batch $d$}
        \STATE forward pass with $[w_0, \ldots, w_{n(w)}]$.
        \STATE backward pass and update $w$.
    \ENDFOR
\ENDFOR
\STATE $m_i \gets \|w_i\|_1$.
\STATE $k \gets \lceil (1 - r) \times n(w) \rceil$.
\FOR{$e \in [s, s+1, \ldots, l]$}
    \FOR{each mini-batch}
        \STATE binary search $t(m_1, \ldots, m_{n(w)})$, find $t$ such that $\sum_{i=1}^{n(w)} f_{\tau,i}(m) = k$.
        \FOR{$i \in [0, \ldots, n(w)]$}
            \STATE $\hat{w}_i \gets w_i \odot f_\tau(m)$.
            \STATE forward pass with $\hat{w}_i$.
        \ENDFOR
        \STATE backward pass and update $w$ and $m$.
        \STATE update $\tau \gets g(\tau_s, \tau_e, e, s, l)$.
    \ENDFOR
\ENDFOR
\STATE $\tilde{m} \gets \text{Top}_k(f_{\tau}(m))$ where $\tilde{m}_i \in \{0,1\}$.
\FOR{$e \in [l + 1, \ldots]$}
    \FOR{each mini-batch}
        \STATE forward pass with $[w_0 \odot \tilde{m}_0, \ldots, w_{n(w)} \odot \tilde{m}_{n(w)}]$.
        \STATE backward pass and update $w$.
    \ENDFOR
\ENDFOR
\end{algorithmic}
\end{algorithm} 

In Algorithm 1, $s$ denotes the total number of epochs allocated for pretraining, while $l$ represents the combined total number of epochs dedicated to both pretraining and structural searching. Additionally, $\tau_s$ and $\tau_e$ signify the starting and ending values of the temperature parameter, respectively. Algorithm 1 outlines a three-phase process in our algorithm: (1) The pre-training stage, which focuses on acquiring a pre-trained model; (2) The structural-searching stage, dedicated to identifying the optimal pruning structure; and (3) The fine-tuning stage, aimed at further improving model performance. 

\subsection{Dynamic temperature parameter trick}
The gradients of the original weights and masks in the SMART pruning algorithm are as follows:
\begin{equation}
\begin{aligned}
\frac{d L}{d w} &= \frac{d L}{d \hat{w}_\tau} \odot f_\tau(m) \\
\frac{d L}{d m} &= \frac{d L}{d \hat{w}_\tau} \odot \left[ \frac{d f_\tau(m)}{d m} w \right] \\
\text{Subject to } \hat{w}_\tau &= f_\tau(m) \odot w
\end{aligned}
\end{equation}
To illustrate the effect of fixed temperature parameters, we introduce Theorem 3.5.
\begin{theorem}
For any given \( \tau \), if the gradients of masked weights equal zero, \( \frac{d L}{d \hat{w}_\tau} = 0 \), then the gradients of original weights \( \frac{d L}{d w} \) and the gradients of mask \( \frac{d L}{d m} \) equals zero.
\end{theorem}

The proof of Theorem 3.5 is straightforward if the feasible region of $\hat{w}_\tau$ is $\mathbb{R}^{N_w}$, with $N_w$ representing the total number of weight elements. As shown in the Equation (6), for any $\hat{w}_\tau \in \mathbb{R}^{N_w}$, there exists a corresponding set of parameters $(m,w)$ that satisfies the constraint $\hat{w}_\tau=f_\tau(m)\odot w$. This confirms that the feasible region for $\hat{w}_\tau$ is indeed $\mathbb{R}^{N_w}$, thereby validating Theorem 3.5. This theorem implies that when the temperature parameter is fixed, our algorithm may converge to a non-sparse local minimum before weights are frozen. This finding is consistent with our ablation study results, which show that a fixed temperature leads to worse performance.

To address this challenge, we integrate a dynamic temperature parameter trick in our approach. This technique entails the continual reduction of the temperature parameter within each mini-batch during the structural searching phase. To demonstrate the efficacy of the dynamic temperature parameter trick in facilitating the escape from non-sparse local minima, we introduce Proposition 3.6. 

\begin{figure}
\centering
\includegraphics[width=0.5\textwidth]{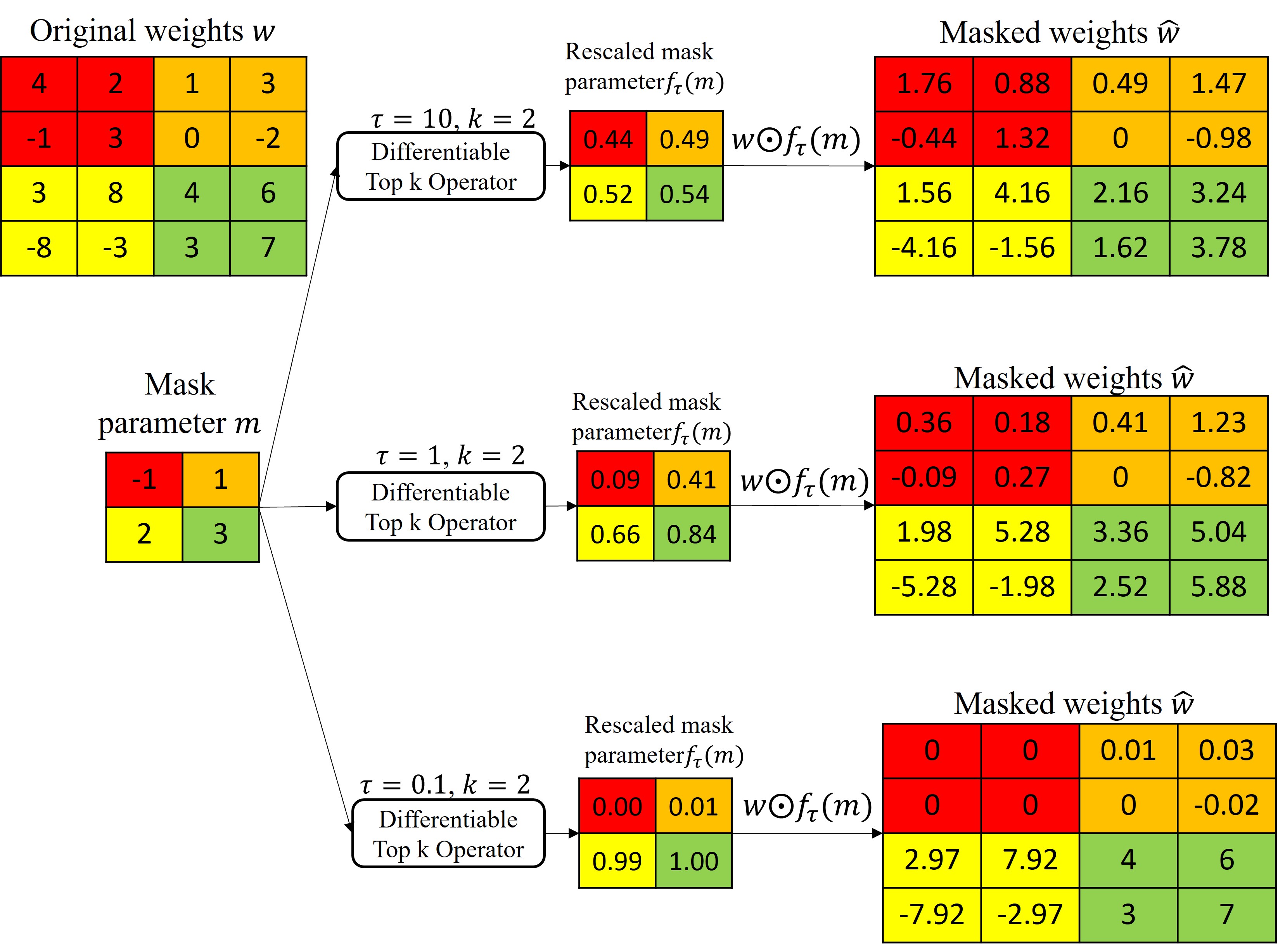} 
\caption{Illustration of the impact of the temperature parameter, $\tau$, on the SMART pruner. As $\tau$ decreases, the rescaled mask parameter, derived from the differentiable Top $k$ operator, more closely approximates binary values (0 or 1), promoting sparsity in the masked weights matrix.}
\label{fig:figure3} 
\end{figure}

\begin{proposition}
For any given \( \tau \) and weights block \( i \), we have
\begin{equation*}
\begin{split}
\lim_{\tau^{*} \to 0} \left\| \hat{w}_{\tau,i} - \hat{w}_{\tau^{*},i} \right\| &= 
\lim_{\tau^{*} \to 0} \left\| \hat{w}_{\tau,i} \right\| \times \left| 1 - \frac{f_{\tau^{*},i}(m)}{f_{\tau,i}(m)} \right| \\
&= \left\| \hat{w}_{\tau,i} \right\| \times \min \left( 1, \left| 1 - \frac{1}{f_{\tau,i}(m)} \right| \right)
\end{split}
\end{equation*}
\end{proposition}
The proof of proposition 3.6 is straightforward. It shows that $ \left\|\hat{w}_{\tau,i} - \hat{w}_{\tau^{*},i} \right\|$ approaches zero if and only if either $\left\|\hat{w}_{\tau,i}\right\|$ is close to zero or $f_{\tau,i}(m)$ is close to one.
In the non-sparse local minimum case, where neither $\left\|\hat{w}_{\tau,i}\right\|$  is close to zero nor $f_{\tau,i}(m)$ is close to one, a notable fluctuation of masked weights occurs due to temperature reduction, as summarized in Table 1. This fluctuation provides an opportunity to escape the non-sparse local minimum, thereby facilitating the continued convergence towards sparse solutions. 

\begin{table} [ht]
\centering
\caption{Fluctuations induced by reducing temperature parameters.}
\label{tab:fluctuations}
\begin{tabular}{@{}lcc@{}} 
\toprule
 & \multicolumn{2}{c}{Masked weights} \\
\cmidrule(l){2-3}
Mask value & Not close to zero & Close to zero \\
\midrule
Not close to one & Large & Negligible \\
Close to one & Negligible & Negligible \\
\bottomrule
\end{tabular}
\end{table}

\begin{theorem}

Suppose \( s \) represents the \( s \)-th iteration in training, and let \( l(s) = t(s)\tau(s) \). If the weight parameters \( w \) are bounded, the loss function \( L \) is continuous with finite gradients, and for any \( i \)-th mask value, we have 
\[
\tau(s) \leq \min\left\{\left(m_i(s) + l(s)\right)^2, \left|\frac{m_i(s) + l(s)}{s}\right|\right\},
\]
then for all mask values \( m_i \), there exists a constant \( m_i^*(s) \) such that
\[
\lim_{s \to \infty} m_i(s) = m_i^*(s).
\]

\end{theorem}

The proof of Theorem 3.7, provided in Appendix E, demonstrates that our algorithm will converge as long as the temperature parameter decays sufficiently fast. In practice, we employ the exponential temperature updating function: $\tau(s) = \tau_{s} - 1 + \beta^{s}$, where $\beta$ is chosen such that $\tau_{e} = \tau_{s} - 1 + \beta^{s}$.

\section{Experimental Results}

We compared our SMART pruner with state-of-the-art block pruning methods across various computer vision tasks and models. Specifically, we benchmarked the SMART pruning algorithm against four methods: PDP, ACDC\citep{peste2021ac}, PaS, and AWG, across three major tasks: classification, object detection, and image segmentation. The AWG pruning algorithm, a modified version of an existing algorithm \citep{lin2018accelerating} for industrial application, is discussed in Appendix F. 

For the classification task, we used the CIFAR-10 dataset \citep{krizhevsky2009learning} to evaluate ResNet18 \citep{he2016deep}, DenseNet \citep{huang2017densely}, GoogleNet \citep{szegedy2015going}, and MobileNetv2 \citep{sandler2018mobilenetv2} models, and the ImageNet dataset \citep{deng2009imagenet} to evaluate ResNet50, with Top-1 accuracy as the performance metric. For object detection, we evaluated YOLOv5m (ReLU version) \citep{jocher2020ultralytics} on the COCO dataset \citep{lin2014microsoft}, using Mean Average Precision (MAP) as the performance metric. For image segmentation, we employed BiSeNetv2 \citep{yu2021bisenet} on the Cityscapes dataset \citep{cordts2015cityscapes}, evaluated with Mean Intersection over Union (MIOU). We compared the performance of different methods across three levels of sparsity: 93\%, 95\%, and 97\% for CIFAR-10 datasets, and 30\%, 50\%, and 70\% for the other tasks and datasets.

Considering the nature of our hardware, we defined the block shape in our study as 16×8×1×1, where 16 represents the number of output channels, 8 represents the number of input channels, and 1×1 are the convolution kernel size dimensions. To compare the performance of our model across different block sizes, we also implemented block pruning with block sizes of 8×8×1×1 and 16×16×1×1 on the CIFAR-10 dataset using ResNet18, DenseNet, GoogleNet, and MobileNetv2 models. To the best of our knowledge, there is no existing research providing tuning parameters for block pruning with these block shapes. Therefore, all models were trained from scratch, and we fine-tuned the competing methods as thoroughly as possible to optimize their performance. Detailed hyper-parameters are available in Appendix G.

\begin{table}[H]
\centering
\small
\caption{The SMART pruning algorithm compared with other benchmark methods on CIFAR-10.}
\begin{subtable}[t]{\columnwidth}
    \centering
    \caption{Block sizes 8x8x1x1 (Top-1 Acc)}
    \resizebox{\columnwidth}{!}{
    \begin{tabular}{@{}cccccc@{}}
    \toprule
    Sparsity & Method & ResNet18 & DenseNet & GoogLeNet & MobileNetv2 \\ 
    \midrule
    Original & - & 0.956 & 0.948 & 0.953 & 0.925 \\ 
    \midrule
    \multirow{6}{*}{93\%} & SMART & \textbf{0.952} & \textbf{0.922} & 0.937 & \textbf{0.921} \\
    & PDP & 0.943 & 0.901 & \textbf{0.939} & 0.919 \\
    & PaS & 0.935 & 0.888 & N/A & 0.903 \\
    & AWG & 0.943 & 0.876 & 0.926 & 0.918 \\
    & ACDC & 0.944 & 0.893 & 0.934 & 0.920 \\
    \midrule
    \multirow{6}{*}{95\%}& SMART & \textbf{0.946} & \textbf{0.899} & \textbf{0.930} & \textbf{0.917} \\
    & PDP & 0.938 & 0.867 & \textbf{0.930} & 0.914 \\
    & PaS & 0.926 & 0.877 & N/A & N/A \\
    & AWG & 0.935 & 0.828 & 0.922 & 0.912 \\
    & ACDC & 0.939 & 0.864 & 0.926 & 0.909 \\
    \midrule
    \multirow{6}{*}{97\%} & SMART & \textbf{0.935} & \textbf{0.850} & \textbf{0.924} & \textbf{0.904} \\
    & PDP & 0.925 & 0.785 & 0.913 & 0.893 \\
    & PaS & N/A & N/A & 0.915 & N/A \\
    & AWG & 0.920 & 0.745 & 0.897 & 0.881 \\
    & ACDC & 0.923 & 0.824 & 0.910 & 0.887 \\
    \bottomrule

    \end{tabular}
}
\end{subtable}

\begin{subtable}[t]{\columnwidth}
    \centering
    \caption{Block sizes 16x8x1x1 (Top-1 Acc)}
    \resizebox{\columnwidth}{!}{
    \begin{tabular}{@{}cccccc@{}}
    \toprule
    Sparsity & Method & ResNet18 & DenseNet & GoogLeNet & MobileNetv2 \\ 
    \midrule
    Original & - & 0.956 & 0.948 & 0.953 & 0.925 \\ 
    \midrule
    \multirow{6}{*}{93\%} & SMART & \textbf{0.951} & \textbf{0.914} & \textbf{0.936} & \textbf{0.917} \\
    & PDP & 0.942 & 0.870 & 0.926 & \textbf{0.917} \\
    & PaS & 0.930 & N/A & 0.930 & N/A \\
    & AWG & 0.939 & 0.866 & 0.922 & 0.913 \\
    & ACDC & 0.940 & 0.884 & 0.929 & \textbf{0.917} \\
    \midrule
    \multirow{6}{*}{95\%}& SMART & \textbf{0.942} & \textbf{0.898} & \textbf{0.929} & \textbf{0.913} \\
    & PDP & 0.933 & 0.850 & 0.920 & 0.910 \\
    & PaS & 0.928 & 0.859 & 0.921 & N/A \\
    & AWG & 0.928 & 0.824 & 0.907 & 0.900 \\
    & ACDC & 0.932 & 0.862 & 0.919 & 0.902 \\
    \midrule
    \multirow{6}{*}{97\%} & SMART & \textbf{0.923} & \textbf{0.844} & \textbf{0.927} & \textbf{0.896} \\
    & PDP & 0.918 & 0.758 & 0.917 & 0.875 \\
    & PaS & 0.921 & N/A & N/A & N/A \\
    & AWG & 0.912 & 0.100 & 0.880 & 0.840 \\
    & ACDC & 0.917 & 0.100 & 0.900 & 0.850 \\
    \bottomrule

    \end{tabular}
}
\end{subtable}
\begin{subtable}[t]{\columnwidth}
    \centering
    \caption{Block sizes 16x16x1x1 (Top-1 Acc)}
    \resizebox{\columnwidth}{!}{
    \begin{tabular}{@{}cccccc@{}}
    \toprule
    Sparsity & Method & ResNet18 & DenseNet & GoogLeNet & MobileNetv2 \\ 
    \midrule
    Original & - & 0.956 & 0.948 & 0.953 & 0.925 \\ 
    \midrule
    \multirow{6}{*}{93\%} & SMART & \textbf{0.946} & \textbf{0.914} & \textbf{0.938} & \textbf{0.914} \\
    & PDP & 0.933 & 0.842 & 0.890 & 0.911 \\
    & PaS & 0.930 & 0.879 & 0.929 & N/A \\
    & AWG & 0.934 & 0.857 & 0.916 & 0.911 \\
    & ACDC & 0.939 & 0.879 & 0.929 & 0.908 \\
    \midrule
    \multirow{6}{*}{95\%}& SMART & \textbf{0.944} & \textbf{0.893} & \textbf{0.930} & \textbf{0.904} \\
    & PDP & 0.925 & 0.707 & 0.856 & 0.893 \\
    & PaS & N/A & 0.811 & N/A & N/A \\
    & AWG & 0.929 & 0.795 & 0.902 & 0.902 \\
    & ACDC & 0.932 & 0.839 & 0.923 & 0.901 \\
    \midrule
    \multirow{6}{*}{97\%} & SMART & \textbf{0.930} & \textbf{0.842} & \textbf{0.912} & \textbf{0.888} \\
    & PDP & 0.908 & 0.594 & 0.818 & 0.858 \\
    & PaS & N/A & 0.815 & 0.901 & N/A \\
    & AWG & 0.911 & 0.100 & 0.873 & 0.836 \\
    & ACDC & 0.920 & 0.100 & 0.898 & 0.843 \\
    \bottomrule

    \end{tabular}
}
\end{subtable}

\end{table}
Our experimental results are shown in Tables 2 (a), (b), (c), and Table 3. In Tables 2 (a)–(c), SMART consistently achieves the best or near-best performance on CIFAR-10 across various block sizes, sparsity levels, and models. Table 3 further demonstrates that SMART outperforms all benchmark methods at every sparsity level across all models and tasks, demonstrating its state-of-the-art status in block pruning applications.

\section{Ablation Study}

In this section, we systematically evaluate the impact of various components and hyperparameters on the performance of the SMART pruning algorithm. We conducted our experiments on the Yolov5 model with $50\%$ sparsity, using MAP as the performance metric. The basic settings are summarized in Table 4, with a MAP of $0.417$ under the default configuration. To better understand the effects of each component, we varied one parameter at a time while keeping the others constant. The detailed results of these changes are summarized in Table 5.

\begin{table}[ht]
\centering
\small
\caption{The SMART pruning algorithm compared with other benchmark methods on other datasets and tasks.}
\resizebox{\columnwidth}{!}{
\begin{tabular}{@{}cccccc@{}}
\toprule
Sparsity & Method & Yolov5m (MAP) & ResNet50 (Top-1 Acc) & BiSeNetv2 (MIOU)  \\
\midrule
Original & - & 0.426 & 0.803 & 0.749\\
\midrule
  \multirow{5}{*}{30\%} & SMART & \textbf{0.424} & \textbf{0.798} & \textbf{0.742}  \\
 & PDP & 0.401 & 0.796 & 0.741  \\
 & PaS & 0.416 & 0.795 & 0.721  \\
 & AWG & 0.404 & 0.793 & 0.741  \\
 & ACDC & 0.403 & 0.795 & \textbf{0.742}  \\
\midrule
 \multirow{5}{*}{50\%}& SMART & \textbf{0.417} & \textbf{0.790} & \textbf{0.736}  \\
 & PDP & 0.384 & 0.787 & 0.728 \\
  & PaS & 0.407 & 0.784 & 0.701  \\
 & AWG & 0.392 & 0.781 & 0.735  \\
 & ACDC & 0.379 & 0.785 & 0.735  \\
\midrule
 \multirow{5}{*}{70\%}& SMART & \textbf{0.398} & \textbf{0.775} & \textbf{0.712}  \\
 & PDP & 0.323 & 0.758 & 0.696  \\
 & PaS & 0.377 & 0.761 & 0.685  \\
 & AWG & 0.355 & 0.758 & 0.691  \\
 & ACDC & 0.318 & 0.759 & 0.705  \\
\bottomrule
\end{tabular}
}
\end{table}

\begin{table} [ht]
\centering
\small
\caption{Ablation Study of the SMART Pruning Algorithm: Basic Settings}
\resizebox{\columnwidth}{!}{

\begin{tabular}{@{}lc@{}}
\toprule
\textbf{Parameter} & \textbf{Value} \\ 
\midrule
Mask Type & Deterministic Probability Masks \\ 
Fixed Temperature & NA \\ 
Dynamic Function Type & Exponential \\ 
Search Iteration & $3.5 \times 10^{4}$ \\ 
Initial Values & 0.5 \\ 
End Values & 0.00001 \\ 
Mask Initialization Method & Mean of absolute weights in the block \\ 
Weight-Frozen Strategy & Unfrozen \\ 
\bottomrule
\end{tabular}
}
\end{table}

\textbf{Mask Types:} We investigated different types of masks to determine their influence on performance. Specifically, we tested three types: deterministic probability masks, which are used in our SMART pruning algorithm; stochastic probability masks, which are utilized in SCP and DSA; and binary masks, employed in DNS. Our results indicate that the deterministic probability mask demonstrated the highest effectiveness, surpassing both the binary mask and the stochastic probability mask. 

\textbf{Temperature Parameter Setting:} The impact of various temperature parameter settings on the performance of the SMART pruning algorithm was thoroughly analyzed. We considered two primary types of temperature settings: fixed temperature parameters and dynamic temperature parameters. For the dynamic temperature parameters, we examined four distinct hyperparameters: dynamic function type, search iterations, initial temperature values, and end temperature values. 

Suppose $\tau(0)$ represents the initial temperature value, $\tau(se)$ represents the end temperature value, and $se$ stands for the search epoch. The temperature schedules are defined as follows: (1) Linear function type: $\tau(n) = \tau(0) - n \frac{\tau(0) - \tau(se)}{se}$. (2) Exponential function type: $\tau(n) = \tau(0) - 1 + \beta^{n}$, where $\beta$ is chosen such that $\tau(se) = \tau(0) - 1 + \beta^{se}$. (3) Inverse exponential function type: $\tau(n) = \tau(0) + 1 - \beta^{n}$, with $\beta$ selected to satisfy $\tau(se) = \tau(0) + 1 - \beta^{se}$.

Our findings are as follows: (1) The dynamic temperature parameter trick is essential, as all fixed temperature settings yielded relatively poor results. (2) The dynamic function type significantly affects the performance of SMART pruning, with the exponential function showing a considerable advantage over linear and inverse exponential functions. (3) Performance remains relatively stable across different search epoches and initial values. (4) Performance is relatively stable when the end values are sufficiently small, but excessively large end values lead to a significant performance drop.

\textbf{Mask Initialization Methods:} We tested two different mask initialization methods: (1) Initialized with the mean of absolute weights in the block, and (2) Initialized with all ones. From our ablation study results, we observed that the SMART pruning algorithm is relatively stable to the mask initialization methods.

\textbf{Weight-Frozen Strategies:} Finally, we explored different strategies for weight freezing, which refers to whether we update weights together with mask training. From the ablation study results, we observed that whether weights were frozen or unfrozen had relatively similar performance.

Our ablation study analyzed how different components and hyperparameter settings impact the final performance of our SMART pruning algorithm. From this analysis, we found that selecting exponential dynamic temperature parameters and sufficiently small end values leads to stable and high performance for the Yolov5 model, even with variations in other parameter settings. Although not included in the ablation studies, we also observed similar stability of the hyperparameters for other models in our experiments. These findings demonstrate the relatively low tuning efforts required for the SMART pruning algorithm, making it suitable for industrial applications.

\begin{table} 
\centering
\small
\caption{Ablation Study of the SMART Pruning Algorithm: Changes in Settings}
\resizebox{\columnwidth}{!}{
\begin{tabular}{@{}lcc@{}}
\toprule
\textbf{Changed Parameter} & \textbf{Value} & \textbf{Performance (MAP)} \\ 
\midrule
Mask Type & Binary Mask & 0.326 \\ 
Mask Type & Stochastic Probability Mask & 0.413 \\ 
\midrule
Fixed Temperature & 0.01 & 0.391 \\ 
Fixed Temperature & 0.001 & 0.398 \\ 
Fixed Temperature & 0.0001 & 0.323 \\ 
Fixed Temperature & 0.00001 & 0.322 \\ 
Fixed Temperature & 0.000001 & 0.322 \\ 
\midrule
Dynamic Function Type & Linear & 0.407 \\ 
Dynamic Function Type & Inverse Exponential & 0.406 \\ 
\midrule
Search Iteration & 5000 & 0.414 \\ 
Search Iteration & 10000 & 0.415 \\ 
Search Iteration & 70000 & 0.416 \\ 
Search Iteration & 140000 & 0.415 \\ 
\midrule
Initial Values & 0.1 & 0.416 \\ 
Initial Values & 1 & 0.416 \\ 
Initial Values & 5 & 0.415 \\ 
Initial Values & 10 & 0.417 \\ 
\midrule
End Values & 0.001 & 0.376 \\ 
End Values & 0.0001 & 0.416 \\ 
End Values & 0.000001 & 0.416 \\ 
End Values & 0.0000001 & 0.415 \\ 
\midrule
Mask Initialization Method & All ones & 0.416 \\ 
\midrule
Weight-Frozen Strategy & Frozen & 0.415 \\ 
\bottomrule
\end{tabular}
}
\end{table}

\section{Conclusion}

In this paper, we present a novel SMART pruning algorithm designed for block pruning applications. SMART addresses key limitations of traditional gradient-based pruners by overcoming three critical challenges: (1) maintaining high accuracy across different models and tasks, (2) efficiently controlling resource constraints, and (3) ensuring convergence guarantees. Our theoretical analysis proves that SMART will converge when the dynamic temperature decays sufficiently fast. Additionally, we show that as the temperature approaches zero, the global optimum of SMART aligns with that of the original pruning problem. Empirical results demonstrate that SMART outperforms existing methods across a variety of computer vision tasks and models. Ablation studies further indicate that using an exponential dynamic temperature function, with a carefully chosen small end temperature, leads to stable performance, reducing the need for extensive parameter tuning and enhancing the practicality of SMART in industrial applications.


\newpage
{\small

\sloppy
\bibliographystyle{apalike}
\bibliography{SMART_Pruning_Paper}
}

\appendix
\onecolumn

\section{Proof of Theorem 3.1}
\begin{proof}
Given the monotonicity of the sigmoid function, we have \(f_{\tau,\rho_i}(x) \leq f_{\tau,\rho_{i+1}}(x); 1 \leq i \leq N - 1\). We prove Theorem 3.1 by contradiction.

Define: \( z_i = \frac{x_{\rho_i}}{\tau} + t(x_1, \ldots, x_N)\), we have (1) \(\lim_{\tau \to 0} \min(z_{i+1} - z_i) = \infty\), (2) \(\sum_{i=1}^{N} \sigma(z_i) = k\), (3) \(0 < \sigma(z_i) < 1\).

Suppose \(\lim_{\tau \to 0} \sigma(z_{N-k}) = a\), where \(0 < a \leq 1\), we have: \(\lim_{\tau \to 0} z_{N-k} \neq -\infty\). Define \(\Delta = z_{N-k+1} - z_{N-k}\), we have: \(\lim_{\tau \to 0} \sigma(z_{N-k+1}) = \lim_{\tau \to 0} \sigma(z_{N-k} + \Delta) = 1\). Given the monotonicity of the sigmoid function, we have:
$\lim_{\tau \to 0} \sum_{i=1}^{N} \sigma(z_i) \geq k + a \neq k$, which is a contradict. Thus, \(\lim_{\tau \to 0} \sigma(z_{N-k}) = 0\).

Similarly, suppose \(\lim_{\tau \to 0} \sigma(z_{N-k+1}) = b\), where \(0 \leq b < 1\), we have: \(\lim_{\tau \to 0} z_{N-k+1} \neq \infty\). Define \(\Delta = z_{N-k+1} - z_{N-k}\), we have: \(\lim_{\tau \to 0} \sigma(z_{N-k}) = \lim_{\tau \to 0} \sigma(z_{N-k+1} - \Delta) = 0\). Given the monotonicity of the sigmoid function, we have:
$\lim_{\tau \to 0} \sum_{i=1}^{N} \sigma(z_i) \leq k - 1 + b \neq k$, which is a contradict. Thus, \(\lim_{\tau \to 0} \sigma(z_{N-k+1}) = 1\).\\
\end{proof}
\section{Proof of Proposition 3.2}
\begin{proof}
The gradient of \( f_i(x) \) can be calculated as follows:
\begin{equation}
\frac{d f_i(x)}{d x_j} = \frac{d \sigma\left(\frac{x_{i}}{\tau} + t(x_1, \ldots, x_N)\right)}{d x_{i}} = \frac{1}{\tau} \left(\frac{x_{i}}{\tau} + t(x_1, \ldots, x_N)\right) \left( I_{\{i=j\}} + \frac{d t(x_1, \ldots, x_N)}{d x_j} \right)
\end{equation}
To obtain the gradient of \( f_i(x) \), we need to determine \( \frac{d t(x_1, \ldots, x_N)}{d x_j} \). To determine the value of \( \frac{d t(x_1, \ldots, x_N)}{d x_j} \), we employ a methodology like the one described in \cite{ahle2019differentiable}. Specifically, the steps for derivation are outlined as follows:
\begin{equation}
0 = \frac{d k}{d x_j} = \frac{d \sum_{i=1}^{N} f_i(x)}{d x_j} = \sum_{i=1}^{N} \frac{1}{\tau} \sigma'\left(\frac{x_{i}}{\tau} + t(x_1, \ldots, x_N)\right) \left( I_{\{j=i\}} + \frac{d t(x_1, \ldots, x_N)}{d x_j} \right)
\end{equation}

Then we have:
\begin{equation}
\frac{d t(x_1, \ldots, x_N)}{d x_j} = -\frac{\sigma'\left(\frac{x_{j}}{\tau} + t(x_1, \ldots, x_N)\right)}{\sum_{i=1}^{N} \sigma'\left(\frac{x_{i}}{\tau} + t(x_1, \ldots, x_N)\right)}
\end{equation}

By plugging Equation (9) into Equation (8), we obtain the following result:
\begin{equation}
\frac{d f_i(x)}{d x_j} = \frac{1}{\tau} \sigma'\left(\frac{x_{i}}{\tau} + t(x_1, \ldots, x_N)\right) \left( I_{\{j=i\}} - \frac{\sigma'\left(\frac{x_{j}}{\tau} + t(x_1, \ldots, x_N)\right)}{\sum_{i=1}^{N} \sigma'\left(\frac{x_{i}}{\tau} + t(x_1, \ldots, x_N)\right)} \right)
\end{equation}

Then, by defining \( v = \left[\sigma'\left(\frac{x_{1}}{\tau} + t(x_1, \ldots, x_N)\right), \ldots, \sigma'\left(\frac{x_{N}}{\tau} + t(x_1, \ldots, x_N)\right)\right]^T \), it follows directly from Equation (10) that \( J_{\text{Top}_k}(x) = \frac{1}{\tau} \left( \text{diag}(v) - vv^T \right) \).
\end{proof}
\section{Proof of Theorem 3.3}
\begin{proof}
We proceed with a proof by contradiction. Assume the existence of a vector $\hat{w}^{**}$
such that $L(\hat{w}^{**}) < L(w^{*}\odot Top_k(m^{*}))$, subject to sparsity constraint $\| \hat{w}^{*}\|_0 = (1-r) \times n(w)$. Let us define $w^{**}=\hat{w}^{**}$ and $m^{**}=\|\hat{w}^{**}\|_1$, we have $ L(w^{**}\odot Top_k(m^{**})) = L(\hat{w}^{**})  < L(w^{*}\odot Top_k(m^{*}))$, which contradicts the premise that the pair $(w^{*},m^{*})$ is the global optimum solution.
\end{proof}
\section{Proof of Theorem 3.4}
\begin{proof}
Suppose \( K_L \) is the Lipschitz constant of loss function \( L \), and define \( \Delta_L = L(\tilde{w} \odot \text{Top}_k(\tilde{m})) - L(w^* \odot \text{Top}_k(m^*)) \), we have:
\begin{align*}
L(\tilde{w} \odot f_\tau(\tilde{m})) - L(w^* \odot f_\tau(m^*)) &= L(\tilde{w} \odot f_\tau(\tilde{m})) - L(\tilde{w} \odot \text{Top}_k(\tilde{m})) \\
&\quad + L(\tilde{w} \odot \text{Top}_k(\tilde{m})) - L(w^* \odot \text{Top}_k(m^*)) \\
&\quad + L(w^* \odot \text{Top}_k(m^*)) - L(w^* \odot f_\tau(m^*)).
\end{align*}

Since \( L \) is a Lipschitz continuous function with Lipschitz constant \( K_L \), we have:
\begin{align*}
L(\tilde{w} \odot f_\tau(\tilde{m})) - L(\tilde{w} \odot \text{Top}_k(\tilde{m})) &\geq -K_L \|\tilde{w} \odot f_\tau(\tilde{m}) - \tilde{w} \odot \text{Top}_k(\tilde{m})\| \\
L(w^* \odot \text{Top}_k(m^*)) - L(w^* \odot f_\tau(m^*)) &\geq -K_L \|w^* \odot f_\tau(m^*) - w^* \odot \text{Top}_k(m^*)\|.
\end{align*}

Then we have:
\begin{align*}
L(\tilde{w} \odot f_\tau(\tilde{m})) - L(w^* \odot f_\tau(m^*)) &\geq \Delta_L - K_L(\|\tilde{w} \odot f_\tau(\tilde{m}) - \tilde{w} \odot \text{Top}_k(\tilde{m})\| + \|w^* \odot f_\tau(m^*) - w^* \odot \text{Top}_k(m^*)\|).
\end{align*}

Theorem 3.1 shows that the differentiable Top k function $f_\tau(m)$ can approximate the standard Top k function $\text{Top}_k(m)$ with arbitrary precision. Thus,
there exists a \( \tau \) such that for any \( \tau \in (0, \tilde{\tau}] \), we have:
\[
\|\tilde{w} \odot f_\tau(\tilde{m}) - \tilde{w} \odot \text{Top}_k(\tilde{m})\| + \|w^* \odot f_\tau(m^*) - w^* \odot \text{Top}_k(m^*)\| < \frac{\Delta_L}{K_L}.
\]

This inequality implies:
\[
L(\tilde{w} \odot f_\tau(\tilde{m})) - L(w^* \odot f_\tau(m^*)) \geq \Delta_L - \Delta_L = 0.
\]
\end{proof}

\section{Proof of Theorem 3.7}

\begin{proof}

Given \( m_i(s+1) = m_i(s) - lr \cdot \frac{dL}{dm_i(s)} \), according to Cauchy's Theorem, proving \( \lim_{s \to \infty} m_i(s) = m_i^* \) is equivalent to proving the following claim: 

For any given \( \epsilon > 0 \), there exists \( N_i \) such that 
\[ 
\sum_{s=N_i+1}^{\infty} \frac{dL}{dm_i(s)}  < \epsilon. 
\]

We have:
\[ 
\frac{dL}{dm_i(s)} = \sum_{j=1}^{n} \frac{dL}{d\hat{w}_j(s)} w_{j} \sigma'\left( \frac{m_j(s) + l(s)}{\tau(s)} \right) \cdot \frac{1}{\tau(s)} \cdot \left[ I_{j=i} - \frac{\sigma'\left( \frac{m_j(s) + l(s)}{\tau(s)} \right)}{\sum_{l=0}^{n} \sigma'\left( \frac{m_j(s) + l(s)}{\tau(s)} \right)} \right]. 
\]

Since the weights are bounded and the loss function is continuous with finite gradients, there exists a finite number \( A \) such that \( \left| \max \left\{ \frac{dL}{d\hat{w}_j(s)} w_{j} \right\} \right| \leq A \) and \( \left| I_{j=i} - \frac{\sigma'\left( \frac{m_j(s) + l(s)}{\tau(s)} \right)}{\sum_{l=0}^{n} \sigma'\left( \frac{m_j(s) + l(s)}{\tau(s)} \right)} \right| < 2 \). Therefore, we obtain:
\[
\begin{aligned}
\frac{dL}{dm_i(s)} &\leq 2A \sum_{j=1}^{n} \sigma'\left( d_j(s) \right) \cdot \frac{d_j(s)}{m_j(s) + l(s)} \\
&= 2A \sum_{j=1}^{n} \frac{d_j(s)}{\exp(d_j(s)) + \exp(-d_j(s)) + 2} \cdot \frac{1}{m_j(s) + l(s)} \\
&\leq 2A \sum_{j=1}^{n} \left| \frac{d_j(s)}{\exp(d_j(s)) + \exp(-d_j(s)) + 2} \right| \cdot \left| \frac{1}{m_j(s) + l(s)} \right|.
\end{aligned}
\]

From Theorem 3.1, it is straightforward to see \( \lim_{s \to \infty} |d_j(s)| = \infty \). Therefore, for sufficiently large \( s \), we have:
\[ 
\sum_{j=1}^{n} \left| \frac{d_j(s)}{\exp(d_j(s)) + \exp(-d_j(s)) + 2} \right| < \sum_{j=1}^{n} 2 \frac{|d_j(s)|}{\exp(|d_j(s)|)}. 
\]

Thus, we obtain:
\[ 
 \frac{dL}{dm_i(s)}  < 4A \sum_{j=1}^{n} \frac{|d_j(s)|}{\exp(|d_j(s)|)} \cdot \left|\frac{1}{m_j(s) + l(s)}\right|.
\]

Given the condition \( \tau(s) \leq \min_j (m_j(s) + l(s))^2 \), it follows that \( \left|d_j(s)\right| \geq \frac{1}{|m_j(s) + l(s)|} \). Similarly, under the condition \( \tau(s) \leq \min_j \left|\frac{m_j(s) + l(s)}{s}\right| \), we have \( \left|d_j(s)\right| \geq s \). Considering the monotonic decreasing behavior of the function \( \frac{x^2}{\exp(x)} \) for large values of \( x \), we obtain:

\[ 
\frac{dL}{dm_i(s)} \leq 4A \sum_{j=1}^{n} \frac{\left|d_j(s)\right|}{\exp(\left|d_j(s)\right|)} \left|d_j(s)\right| \leq 4A \sum_{j=1}^{n} \frac{s^2}{\exp(s)} = 4An \frac{s^2}{\exp(s)}.
\]

Since \( \lim_{s \to \infty} \frac{s^2}{\exp(s)} < \frac{1}{\exp(s/2)}\), we have for any given \( \epsilon \), there exists \( N_i \) such that 
\[ 
\sum_{s=N_i + 1}^{\infty}\frac{dL}{dm_i(s)} \leq 4An\sum_{s=N_i + 1}^{\infty} \frac{s^2}{\exp(s)} < 4An\sum_{s=N_i + 1}^{\infty} \frac{1}{\exp(s/2)} = 4An\frac{\exp(-\frac{N_i+1}{2})}{1-\exp(-\frac{1}{2})} < \epsilon.
\]

Therefore, we conclude that 
\[ 
\lim_{s \to \infty} m_i(s) = m_i^*.
\]

\end{proof}
\newpage
\section{The Accumulated Weight and Gradient (AWG) Pruning Algorithm}

The Accumulated Weight and Gradient pruner is an advanced variant of the magnitude-based pruning method, where the importance score is a function of the pre-trained weights and the tracked gradients over one epoch of gradient calibration. As opposed to the pre-trained weights, the product of pre-trained weights and accumulated gradients is assumed to be the proxy for importance. For each layer, the importance score is the product of three terms. 1) The pre-trained weight. 2) The accumulated gradient. 3) The scaling factor, which rescales the importance by favoring (higher score) layers with high sparsity already.

The pruning consists of three stages. 1) Initially, the training data is fed to the pre-trained model and we keep track of the smoothed importance scores via exponential moving average (EMA) at each layer. For more detail regarding how the importance score is computed and updated, please refer to Algorithm 3. Weights are not updated at this stage. After running one epoch, we reduce the importance scores by block and rank them in the ascending order. We prune the least important $p\%$ of the blocks through 0-1 masks. 2) With the masks updated, we fine-tune the weights on the training set for $P$ epochs. The masks are kept unchanged throughout the fine-tuning. Typically, we carry out iterative pruning to allow for smoother growth in sparsity and thus more stable convergence, so stages 1 and 2 alternate for $S$ steps. 3) At the very last step, we fine-tune the model for another $Q$ epochs.

A brief explanation for the notation in the algorithm: $S$ is the number of iterative steps in pruning. $\gamma$ is the decay factor that governs the EMA smoothing for importance. $P$ is the number of epochs to fine-tune the model at each iterative step. $Q$ is the number of epochs to fine-tune the model after the last iterative step is complete. $w_i$, $m_i$ and $imp_i$ are the block-wise weight, mask and EMA importance of the $i$-th block, respectively.

\begin{algorithm}
\small
\caption{Training flow for Accumulated Weight and Gradient pruner}
\begin{algorithmic}
\STATE \textbf{Input:} $S, \gamma, P, Q, r, w = [w_1, \ldots, w_{n(w)}], m = [m_1, \ldots, m_{n(w)}], imp= [imp_1, \ldots, imp_{n(w)}]$.
    \STATE $m \gets 1$
    \FOR{$k \in [1, 2, \ldots, S]$}
        \FOR{first mini-batch}
            \STATE forward pass with $w \odot m$.
            \STATE backward pass to update $w$.
            \STATE $imp_i = \left| grad_{w_i} \odot w_{i} \odot \frac{\left\|m_{J}\right\|_0}{\sum m_{J}}\right|$, where $J$ is the layer-level mask to which $m_i$ belongs.
        \ENDFOR
        \FOR{mini-batch after the first mini-batch}
            \STATE forward pass with $w \odot m$.
            \STATE backward pass to update $w$.
            \STATE $imp_i = \gamma \cdot imp_i + (1-\gamma) \cdot \left| grad_{w_i} \odot w_{i} \odot \frac{\left\|m_{J}\right\|_0}{\sum m_{J}}\right|$.
        \ENDFOR
        \STATE rank the importance in ascending order.
        \STATE compute the threshold $t$ such that the $\left( \frac{r}{S} k \right)$-th quantile of the importance scores falls under $t$.
        \FOR{each block/channel in $m$}
            \STATE $m_i \gets 0$ if $\text{imp}_i < t$, otherwise $m_i \gets 1$.
        \ENDFOR
        \FOR{$e \in [0, P]$}
            \FOR{each mini-batch}
                \STATE forward pass with $w \odot m$.
                \STATE backward pass to update $w$.
            \ENDFOR
        \ENDFOR
        \ENDFOR

        \FOR{$e \in [0, Q]$}
            \FOR{each mini-batch}
                \STATE forward pass with $w \odot m$.
                \STATE backward pass to update $w$.
            \ENDFOR
        \ENDFOR
\end{algorithmic}
\end{algorithm}

\begin{landscape}
    \section{Hyperparameter Settings}
    \small

    \subsection{CIFAR-10}
    \begin{centering}
    \begin{tabular}{p{1.5cm}|p{1cm}|p{1cm}|p{1cm}|p{5cm}|p{3cm}|p{6cm} }
     \hline
     Network & Method & Batch size & Epochs  & Main optimizer; scheduler & Mask optimizer; scheduler & Pruning-specific params \\
     \hline \hline
     \multirow{6}{4em}{ResNet18}
     & AWG & 128 & 200  & SGD: 0.02, 0.9, 5e-4; cosine & N/A & mspl: 0.98 \\
    & ACDC & 128 & 200  & SGD: 0.02, 0.9, 5e-4; cosine & N/A & mspl: 0.98, steps: 8, \#comp: 1955, \#decomp: 1955 \\
    & PaS & 128 & 200  & SGD: 0.005, 0.9, 5e-4; cosine & Same as the main & lambda: See PaS Lambda \\
    & PDP & 128 & 200  & SGD: 0.02, 0.9, 5e-4; cosine & Same as the main & \#si: 15640 \\
    & Ours & 128 & 200  & SGD: 0.02, 0.9, 5e-4; cosine & Same as the main & init\_temp: see notes, final\_temp: 1e-4, \#si: 15640 \\
     \hline
     \multirow{6}{4em}{DenseNet}
    & AWG & 128 & 200  & SGD: 0.02, 0.9, 5e-4; cosine & N/A & mspl: 0.98 \\
    & ACDC & 128 & 200  & SGD: 0.02, 0.9, 5e-4; cosine & N/A & mspl: 0.98, steps: 8, \#comp: 1955, \#decomp: 1955 \\
    & PaS & 128 & 200  & SGD: 0.005, 0.9, 5e-4; cosine & Same as the main & lambda: See PaS Lambda \\
    & PDP & 128 & 200  & SGD: 0.02, 0.9, 5e-4; cosine & Same as the main & \#si: 15640 \\
    & Ours & 128 & 200 & SGD: 0.02, 0.9, 5e-4; cosine & Same as the main & init\_temp: see notes, final\_temp: 1e-4, \#si: 15640 \\
     \hline
     \multirow{6}{4em}{GoogLeNet}
    & AWG & 128 & 200  & SGD: 0.02, 0.9, 5e-4; cosine & N/A & mspl: 0.98 \\
    & ACDC & 128 & 200  & SGD: 0.02, 0.9, 5e-4; cosine & N/A & mspl: 0.98, steps: 8, \#comp: 1955, \#decomp: 1955 \\
    & PaS & 128 & 200  & SGD: 0.005, 0.9, 5e-4; cosine & Same as the main & lambda: See PaS Lambda \\
    & PDP & 128 & 200  & SGD: 0.02, 0.9, 5e-4; cosine & Same as the main & \#si: 15640 \\
    & Ours & 128 & 200  & SGD: 0.02, 0.9, 5e-4; cosine & Same as the main & init\_temp: 0.01, final\_temp: 1e-4, \#si: 7640 \\
     \hline
     \multirow{6}{4em}{MobileNetv2}
    & AWG & 128 & 200  & SGD: 0.02, 0.9, 5e-4; cosine & N/A & mspl: 0.98 \\
    & ACDC & 128 & 200  & SGD: 0.02, 0.9, 5e-4; cosine & N/A & mspl: 0.98, steps: 8, \#comp: 1955, \#decomp: 1955 \\
    & PaS & 128 & 200  & SGD: 0.005, 0.9, 5e-4; cosine & Same as the main & lambda: See PaS Lambda \\
    & PDP & 128 & 200  & SGD: 0.02, 0.9, 5e-4; cosine & Same as the main & \#si: 15640 \\
    & Ours & 128 & 200  & SGD: 0.02, 0.9, 5e-4; cosine & Same as the main & init\_temp: see notes, final\_temp: 1e-4, \#si: 7640 \\
    \hline
    \end{tabular}
    \end{centering}

    \textbf{Learning Rates}
    \begingroup
    \setlength{\parskip}{-0.5em}
    \begin{itemize}
        \item ResNet18, SMART, 16x8x1x1, 0.93, 0.01
        \item ResNet18, SMART, 16x8x1x1, 0.95, 0.005
        \item ResNet18, SMART, 16x8x1x1, 0.97, 0.01
        \item DenseNet, SMART, 16x8x1x1, 0.93, 0.01
        \item DenseNet, SMART, 16x8x1x1, 0.95, 0.01
        \item DenseNet, SMART, 16x8x1x1, 0.97, 0.005
    \end{itemize}
    \endgroup
    
    \textbf{PaS Lambda}
    \begingroup
    \setlength{\parskip}{-0.5em}
    \begin{itemize}
        \item ResNet18: 16x8x1x1: (13.554, 143.17, 204.732), 8x8x1x1: (301.5, 379.78, N/A), 16x16x1x1: (368.23, N/A, N/A)
        \item DenseNet: 16x8x1x1: (N/A, 143.285, N/A), 8x8x1x1: (94.132, 304.01, N/A), 16x16x1x1: (80.979, 80.0, 80.0)
        \item GoogleNet: 16x8x1x1: (244.891, 228.566, N/A), 8x8x1x1: (N/A, N/A, 264.48), 16x16x1x1: (171.426, N/A, 354.125)
        \item MobileNetv2: 16x8x1x1: (N/A, N/A, N/A), 8x8x1x1: (270.264, N/A, N/A), 16x16x1x1: (N/A, N/A, N/A)
    \end{itemize}
    \endgroup

    \textbf{SMART Hyperparams}
    \begingroup
    \setlength{\parskip}{-0.5em}
    \begin{itemize}
        \item ResNet18
            \begin{itemize}
                \item Init\_temp: 16x8x1x1: 0.1, 8x8x1x1: 0.1, 16x16x1x1: 1e-3
                \item \#si: 16x8x1x1: 7640, 8x8x1x1: 7640, 16x16x1x1: 15640
            \end{itemize}
        \item DenseNet
            \begin{itemize}
                \item Init\_temp: 16x8x1x1: 1e-3, 8x8x1x1: 0.1, 16x16x1x1: 0.1
                \item \#si: 16x8x1x1: 15640, 8x8x1x1: 7640, 16x16x1x1: 7640
            \end{itemize}
        \item MobileNetv2
            \begin{itemize}
                \item Init\_temp: 16x8x1x1: (1, 0.1, 0.1), 8x8x1x1: (1, 0.1, 1), 16x16x1x1: (1, 1, 0.1)
            \end{itemize}
    \end{itemize}
    \endgroup

    \textbf{Notes}:
    \begin{itemize}
        \item If an experiment's learning rate differs from the main table, then the learning rate used is listed underneath the table in the form of \textbf{network, method, block size, sparsity, learning rate}.
        \item The hyperparameters are given in the form of
        \begin{itemize}
            \item network: block size: (sparsity 0.93, sparsity 0.95, sparsity 0.97), if different values are used in different sparsities
            \item network: block size: value, if the same value is used across all sparsities
        \end{itemize}
    \end{itemize}

    \subsection{Other Datasets}

    \begin{centering}
    \begin{tabular}{p{1.5cm}|p{1cm}|p{1cm}|p{1.2cm}|p{5cm}|p{3cm}|p{6cm} }
     \hline
     Network (Dataset) & Method & Batch size & Epochs (Iterations)  & Main optimizer; scheduler & Mask optimizer; scheduler & Pruning-specific params \\
     \hline \hline
     \multirow{6}{4em}{ResNet50 (ImageNet)}
     & AWG & 128 & 100  & AdamW: 1e-4, 0.9, 0.025; cosine & N/A & mspl: 0.98 \\
    & ACDC & 256 & 100  & AdamW: 1e-4, 0.9, 0.025; cosine & N/A & mspl: 0.98, steps: 8, \#comp: 25,000, \#decomp: 25,000 \\
    & PaS & 128 & 100  & AdamW: 1e-5, 0.9, 0.025; cosine & Same as the main & N/A \\
    & PDP & 256 & 100  & AdamW: 1e-5, 0.9, 0.025; cosine & Same as the main & \#si: 100,000 \\
    & Ours & 256 & 100  & AdamW: 1e-4, 0.9, 0.025; cosine & Same as the main & init\_temp: 10, final\_temp: 1e-5, \#si: 100,000 \\
     \hline
     \multirow{6}{4em}{YOLO v5 (COCO)}
     & AWG & 16 & 100 & SGD: 1e-4, 0.937, 5e-4; lambdaLR & N/A & mspl: 0.98 \\
    & ACDC & 32 & 100 & SGD: 1e-4, 0.937, 5e-4; lambdaLR & N/A & mspl: 0.98, steps: 8, \#comp: 18,485, \#decomp: 18,485 \\
    & PaS & 16 & 100 & SGD: 1e-4, 0.937, 5e-4; lambdaLR & SGD: 1e-3, 0.9, 5e-4, lambdaLR & N/A \\
    & PDP & 32 & 100 & SGD: 1e-4, 0.937, 5e-4; lambdaLR & Same as the main & \#si: 73,940 \\
    & Ours & 16 & 100 & SGD: 1e-4, 0.937, 5e-4; lambdaLR & Same as the main & init\_temp: 0.5, final\_temp: 1e-5, \#si: 35,000 \\
     \hline
     \multirow{6}{4em}{BiSeNet v2 (CityScapes)}
     & AWG & 16 & (100,000) & SGD: 5e-4, 0.9, 0; WarmupPolyLR & N/A & mspl: 0.98 \\
    & ACDC & 16 & (100,000) & SGD: 5e-4, 0.9, 0; WarmupPolyLR & N/A & mspl: 0.98, steps: 8, \#comp: 10,000, \#decomp: 10,000 \\
    & PaS & 16 & (100,000) & SGD: 5e-4, 0.9, 0; WarmupPolyLR & Same as the main & N/A \\
    & PDP & 16 & (100,000) & SGD: 5e-4, 0.9, 0; WarmupPolyLR & Same as the main & \#si: 52,400 \\
    & Ours & 16 & (100,000) & SGD: 5e-4, 0.9, 0; WarmupPolyLR & SGD: 1e-3, 0.9, 0; WarmupPolyLR & init\_temp: 0.1, final\_temp: 1e-5, \#si: 1,400 \\
    \hline
    \end{tabular}
    \end{centering}    
    
\textbf{Notes:}
\begingroup
\setlength{\parskip}{-0.5em}
\endgroup

Optimizer Parameters:
\begingroup
\setlength{\parskip}{-0.5em}

\begin{itemize}
    \setlength\itemsep{-0.3em}
    \item SGD: lr, momentum, weight\_decay
    \item Adam: lr, weight\_decay
    \item AdamW: lr, momentum, weight\_decay
\end{itemize}
\endgroup

Pruning Parameters:
\begingroup
\setlength{\parskip}{-0.5em}
\begin{itemize}
    \setlength\itemsep{-0.3em}
    \item AWG: mspl = maximum sparsity per layer, steps = iterative steps, \#feps = number of fine-tuning epochs per step
    \item ACDC: mspl = maximum sparsity per layer, steps = iterative steps, \#comp = number of compressed iterations per step, \#decomp = number of decompressed iterations per step
    \item PaS: lambda = the final lambda used. N/A means no lambda is found that would lead to the desired sparsity
    \item PDP: \#si = number of search iterations
    \item Ours (SMART): init\_temp = initial temperature, final\_temp = final temperature, \#si = number of search iterations
\end{itemize}
\endgroup

\end{landscape}

\end{document}